\documentclass[10pt,twocolumn]{article}

\usepackage[margin=0.8in, headsep=40pt]{geometry}
\usepackage{graphicx}
\usepackage{fancyhdr}
\usepackage{url}
\usepackage{hyperref}
\usepackage{amsmath, amssymb, amsfonts}
\usepackage{booktabs}
\usepackage[inkscapelatex=false]{svg}
\fancypagestyle{firstpage}{
  \fancyhf{}
  \fancyhead[L]{\small Gunbir Singh Baveja \\
  \small gbaveja@student.ubc.ca}
  \fancyhead[R]{\includegraphics[height=1cm]{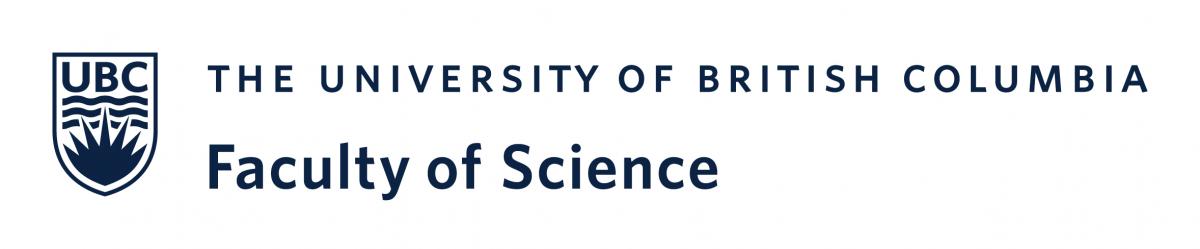}}
}

\title{\textbf{Exploration and Adaptation in Non-Stationary Tasks with Diffusion Policies}}
\author{
    \textbf{Gunbir Singh Baveja}\thanks{The full code is available at \url{https://github.com/sheeerio/continual-diffusion}}\\
    University of British Columbia\\
    Vancouver, Canada
}

\date{}

\begin{document}

\maketitle
\thispagestyle{firstpage}

\section{Introduction and Motivation}

The challenge of achieving robust performance in non-stationary Reinforcement Learning (RL) environments, where the underlying dynamics and objectives evolve over time, encapsulates the complexities of real-world decision-making scenarios. Such tasks often demand that an agent operate directly from high-dimensional visual inputs, a setting that significantly intensifies the learning problem. In many practical applications—ranging from continuously shifting assembly lines in robotics to changing terrain layouts in autonomous navigation—agents must not only learn to adapt control strategies on-the-fly but also maintain long-term stability under ever-changing conditions.

Recent advancements in machine learning have introduced the concept of \textit{Diffusion Policy} \cite{ref1,ref2}, a novel approach that leverages the iterative refinement of actions through a stochastic denoising process \cite{ddpm}. Originally demonstrated in high-dimensional, multimodal robotic control tasks, Diffusion Policy has shown promising capabilities for navigating complex visual environments. By incrementally reducing noise in latent action representations, this generative paradigm aims to produce policies capable of handling intricate observation-to-action mappings.

In personal communications with Dr. Cong Lu, the lead author of the Synthetic Experience Replay approach, it became evident that non-stationary visual decision-making tasks represent a formidable and, in some cases, adversarial challenge for Diffusion Policy. These problems do not merely require the agent to achieve singular, state-specific goals; rather, they demand sustained adaptability, where the policy must continuously respond to evolving conditions. The one-dimensional or low-dimensional action spaces often found in classic benchmarks contrast sharply with the expansive domains where Diffusion Policy has previously excelled, further testing its generalizability and resilience. This aligns with the broader challenges in continual learning \cite{parisi2019continual}, where models must adapt to new tasks without forgetting previously acquired knowledge.

\begin{figure*}[!t]
\centering
\includegraphics[width=\textwidth]{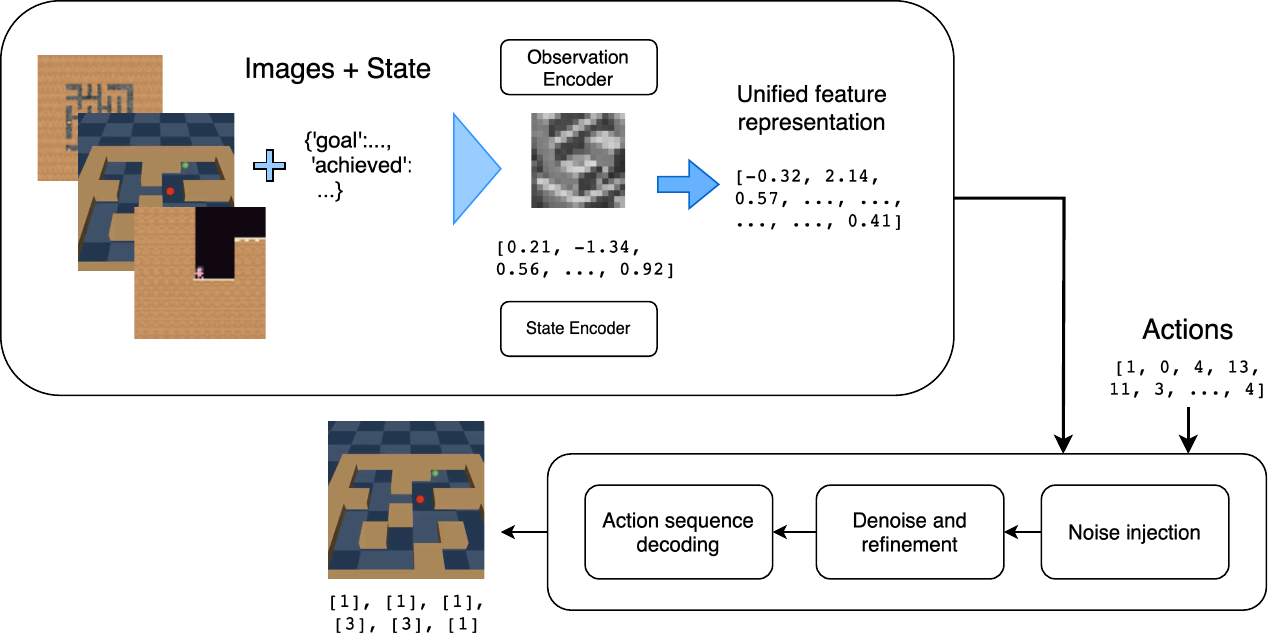}
\caption{\textbf{Unified Representation and Action Decoding in Diffusion Policy.} Visual observations and state information (e.g., goal and achievement metrics) are encoded into a unified feature representation through an observation encoder and state encoder. This representation forms the input to a diffusion-based process, where noise is injected, denoised, and refined to decode action sequences. The framework effectively combines multimodal inputs for robust policy generation in complex navigation tasks.}
\label{fig:problem_setup}
\end{figure*}

Motivated by these insights, this work explores the application of Diffusion Policy to a suite of non-stationary, vision-based RL environments drawn from benchmarks such as Procgen \cite{procgen} and PointMaze \cite{d4rl}. By employing Diffusion Policy in these highly variable settings, we seek to determine whether the generative strengths of this approach can be leveraged to maintain stable and adaptive planning. A visual depiction of the overall policy architecture is shown in Figure~\ref{fig:problem_setup}, highlighting how raw visual observations are transformed into actionable commands through the iterative denoising process.

The experimental results reveal that the Diffusion Policy consistently outperforms traditional RL algorithms like PPO and DQN across multiple tasks, demonstrating superior mean and maximum rewards with lower variability. These findings suggest that the Diffusion Policy's ability to generate coherent and contextually relevant action sequences provides a robust framework for handling non-stationary and dynamic environments. However, challenges such as high computational demands and limitations in handling extreme non-stationarity indicate areas for future improvement and exploration.

\section{Problem Setup and Data Collection}

The problem of controlling agents in non-stationary environments, especially when guided solely by visual inputs, presents a landscape of continuously shifting challenges. In classical stationary benchmarks, the agent is afforded a stable set of dynamics, enabling policy optimization to converge steadily. However, in non-stationary tasks—such as those posed by the Procgen Benchmark \cite{procgen}, where levels and obstacles change across episodes—this stability is lost. The agent must repeatedly reacquire pertinent information about state transitions, underlying dynamics, and reward structures. This fluidity demands that the policy exhibit a form of continual adaptability, rather than relying on a fixed or slowly updating behavior \cite{parisi2019continual}.

In these environments, we consider a variety of tasks that emulate the unpredictable nature of real-world conditions. One such task is the \textbf{CoinRun} environment, where the agent is tasked with navigating procedurally generated platformer-like levels, each featuring novel obstacles, terrain features, and reward placements. The difficulty arises not from a single static challenge but from a stream of unique episodes, each demanding rapid strategic recalibration. Another benchmark environment, the \textbf{Maze} scenario, requires the agent to navigate a diverse array of labyrinths constructed with randomized topologies. Here, the need for spatial understanding, pathfinding strategies, and adaptation to newly generated layouts is paramount.

For continuous planning and more complex physics-based navigation, the \textbf{PointMaze} environment, part of the D4RL suite \cite{d4rl}, offers a complementary challenge. The agent must move a point-mass through static and dynamic maze configurations toward a specified goal. While the layout may remain fixed in some variants, other versions introduce dynamic elements that shift over time, demanding real-time adjustments from the agent. We added this environment as this proves robustness in stationary settings where planning is the main bottleneck.

\subsection{Environment Dynamics}

\textit{Action Space:} CoinRun and Maze feature discrete action spaces, each with 15 possible actions. In CoinRun, these actions include movements such as jumping, moving left, or moving right, enabling efficient navigation through obstacles and successful coin collection. Similarly, Maze actions consist of up, down, left, and right movements, allowing the mouse to navigate the labyrinth in search of the cheese. In contrast, PointMaze operates within a continuous action space, where precise control inputs dictate the point-mass's movement in any direction.

\textit{Observation Space:} The observations in CoinRun and Maze are represented as three-dimensional arrays of shape (64, 64, 3), corresponding to RGB images. In CoinRun, these images provide a visual depiction of the terrain, obstacles, and objectives, while in Maze, they offer a top-down view that highlights paths, walls, and the cheese's location. In PointMaze, the observation space is unified, incorporating both RGB images resized to (64, 64) pixels and a low-dimensional state vector that includes the agent's current position and the locations of the achieved and desired goals.

\textit{Rewards:} The reward structure in all three environments is task-specific and incentivizes goal-oriented behavior. In CoinRun, rewards are provided for successful coin collection, with penalties for falling into chasms or colliding with enemies. Maze rewards are tied to efficiently finding the cheese, penalizing redundant or inefficient movements. PointMaze rewards the agent for reaching the desired goal as quickly and accurately as possible, with the potential for penalties in scenarios where the agent veers off course or fails to maintain efficient navigation.

\textit{Starting State:} Each episode begins with unique initial conditions across all environments. In CoinRun, the agent always starts at the far-left side of the procedurally generated level, whereas in Maze, the mouse and the cheese are randomly placed within the labyrinth, creating a fresh challenge for each episode. PointMaze also features variable initial conditions, with the agent and the goal locations randomized in each episode.

\subsection{Data Collection for Training}

Data collection in these scenarios mirrors the intricacies of the underlying tasks. We must aggregate a diverse dataset $\mathcal{D}$ of state-action-reward sequences that capture both expected and unusual conditions. Early-stage data collection often involves exploratory or even random policies that deliberately seek out new corners of the state space. As we progress, iterative refinement strategies ensure that new trajectories are continuously integrated, enabling the policy to remain sensitive to changes. In discussions with Dr. Cong Lu, who emphasized the importance of representative and continually updated experience, it became evident that carefully curating this growing dataset is crucial. Without a broad and evolving set of trajectories, the generative model would struggle to grasp the full complexity of the environment, ultimately hindering adaptation.

For Procgen environments like CoinRun and Maze, the data generation process involves deploying trained or exploratory policies to collect high-dimensional visual observations paired with discrete actions. This is achieved through a pipeline that leverages convolutional neural networks (CNNs) to process RGB images of shape (64, 64, 3), capturing the rich visual information necessary for navigating procedurally generated levels. Actions are sampled either randomly or based on a pre-trained policy, and the resulting state-action-reward trajectories are stored in a structured format using tools like Zarr \cite{zarr} for efficient access and scalability.

Within each collected trajectory, the policy’s visual inputs—often 2D RGB frames—undergo transformations by a suite of data augmentation techniques. These enhancements serve to enrich the policy’s viewpoint, building invariance to common perturbations such as viewpoint shifts or minor scene variations. Together, these steps form the foundation for a training process that aims to instill in the policy not merely an understanding of a single environment configuration, but a capacity to cope with the broader distribution of conditions that the problem space may produce.

In contrast, the PointMaze environment requires a different data collection strategy due to its continuous action space and the inclusion of state vectors alongside visual inputs. The data generation script initializes the PointMaze environment, captures both the rendered RGB frames resized to (64, 64) pixels and the underlying state information such as the agent's position and goals. A sequence of actions is sampled from a continuous distribution, reflecting a plan needed to navigate the maze.

\section{Model Architecture}

Translating high-dimensional visual inputs into effective planned actions amidst non-stationary dynamics is a non-trivial challenge. Drawing inspiration from the success of diffusion-based generative methods in complex domains \cite{ddpm}, we integrate a \textbf{Denoising Diffusion Probabilistic Model (DDPM)} directly into the reinforcement learning pipeline. The core idea is to iteratively refine noisy, preliminary action sequences into coherent, context-appropriate behaviors. By conditioning this refinement process on latent representations derived from raw images and, when necessary, low-dimensional state vectors, the agent learns to produce robust policies that can respond dynamically to evolving environmental conditions.

\subsection{Visual Encoder and State Representation}

At the forefront of our architecture is a \textit{Visual Encoder}, implemented as a modified ResNet \cite{he2016resnet} pre-trained on a large-scale dataset to leverage a rich, general-purpose feature representation. Additionally, inspired by the U-Net architecture \cite{ronneberger2015u}, we incorporate skip connections to retain spatial information critical for precise action decoding. As shown in the code excerpts, we replace standard BatchNorm layers with GroupNorm to enhance stability when training on smaller or more diverse datasets. The encoder receives image observations as input and outputs a compact visual embedding. In tasks like CoinRun and Maze, where discrete actions and dense visual cues dominate, the encoder’s feature maps capture spatial structures and highlight task-relevant details (e.g., obstacles, collectibles, or maze topologies). For PointMaze, we incorporate state vectors (e.g., agent and goal positions) alongside visual features, ensuring that the agent can handle continuous control scenarios and more physics-based navigational challenges. This unified representation allows the policy to transition seamlessly between discrete and continuous action domains, preserving stability even as the underlying environment or action space changes.

As illustrated in the accompanying code snippet, observation frames are first processed by the vision encoder to extract high-level features. When training on non-stationary tasks, the model can stack multiple frames over an observation horizon $T_o$, thus enabling temporal reasoning. These temporally stacked embeddings provide the DDPM with historical context, helping it infer dynamics like enemy movement in CoinRun or evolving maze layouts in both Maze and PointMaze. This approach parallels strategies used in other diffusion-based studies, where the observation horizon ensures that the model retains essential time-dependent information. Additionally, exploring transformer-based encoders \cite{vaswani2017attention} could further enhance the model's ability to capture long-range dependencies in temporal data.

\subsection{Diffusion Model and Closed-Loop Control}

The Diffusion Model lies at the heart of this policy framework. As outlined in the included code, we employ a Conditional U-Net architecture adapted for 1D temporal sequences of actions. This network receives a noisy action proposal along with a global conditioning vector derived from the visual encoder’s latent features. During training, the DDPM progressively “denoises” action sequences through multiple diffusion steps, ensuring that each iteration refines the action plan according to the observed context. By the final iteration, the resulting action sequence is both feasible and contextually aligned with current environmental conditions.

A key advantage of this diffusion-based design is its capacity for \textbf{closed-loop control}. Rather than committing to an extended open-loop sequence, the agent executes only the first action from the refined proposal at each step, re-encodes the new observation, and re-initiates the diffusion process for the next time step. This iterative feedback loop allows the policy to adapt on-the-fly as new states emerge, providing resilience against non-stationarity. The code snippet demonstrates how actions, states, and rewards are collected and integrated into the training pipeline. Actions are normalized and denormalized to stabilize training, while noise scheduling and loss computation ensure a controlled refinement process that maintains stable gradient flows. Moreover, employing optimizers with decoupled weight decay \cite{loshchilov2017decoupled} like AdamW contributes to more stable and efficient training dynamics.

\section{Training}

\subsection{Diffusion Model Training}

As shown in the provided code, the training procedure commences by loading demonstration datasets stored in Zarr format. These datasets include arrays of images, actions, and other relevant observations collected from various environments. The Diffusion Model is trained by minimizing the mean squared error (MSE) between predicted and true noise components at each diffusion step:
\[
\mathcal{L} = \mathbb{E}_{\mathbf{a}_t, t} \left[\left\|\mathbf{a}_{t-1} - \mu_\theta(\mathbf{a}_t, \phi(s), t) \right\|^2 \right],
\]
where $\phi(s)$ is the latent state embedding from the visual encoder and $\mu_\theta$ represents the denoising predictions. The noise scheduler, implemented via the \texttt{DDPMScheduler}, orchestrates the noise injection and removal process, allowing the model to learn stable and consistent mappings from noisy to clean action sequences. The optimizer (e.g., AdamW) and learning rate scheduler ensure stable and efficient convergence. Additionally, an EMA (Exponential Moving Average) model is maintained to stabilize training and capture a smoothed version of the learned policy.

\subsection{Policy Integration and Fine-Tuning}

Following the initial training phase, the diffusion-based model is integrated into a closed-loop policy. The code snippets demonstrate how new episodes, collected under evolving conditions, are leveraged to iteratively fine-tune the Diffusion Model. By continuously incorporating fresh data and performing updates, the policy remains adaptive and responsive to environmental changes. This iterative refinement mirrors the deployment scenario in non-stationary tasks: as new challenges emerge, the policy’s generative action model is re-calibrated to ensure consistent performance. Additionally, leveraging exploration strategies \cite{bellemare2016unifying} can further enhance the policy's ability to discover effective strategies in diverse environments.

\section{Results}

\subsection{Overview of Evaluation Tables}

For ease of navigation and readability, the evaluation results are presented in two separate tables. Table~\ref{tab:baseline_performance} showcases the baseline performance of the Diffuser compared to PPO and DQN across the CoinRun, Maze, and PointMaze tasks for $500$K steps each. Table~\ref{tab:ablation_studies} details the ablation studies conducted on the Diffuser within the CoinRun environment, examining the effects of different model configurations.

\begin{table*}[t]
\centering
\label{tab:baseline_performance}
\resizebox{350pt}{!}{%
\begin{tabular}{l|rrr|rrr|rrr}
\toprule
\textbf{Task Name} & \multicolumn{3}{c|}{\textbf{Diffuser}} & \multicolumn{3}{c|}{\textbf{PPO}} & \multicolumn{3}{c}{\textbf{DQN}} \\
\midrule
                       & \textbf{Mean} & \textbf{Max} & \textbf{Std} & \textbf{Mean} & \textbf{Max} & \textbf{Std} & \textbf{Mean} & \textbf{Max} & \textbf{Std} \\
\midrule
CoinRun            & 8.15         & 8.30        & 0.15         & 7.95         & 8.10        & 0.22         & 7.60         & 7.80        & 0.28         \\
Maze               & 3.23         & 5.2        & 1.20         & 8.70         & 8.90        & 0.12         & 8.20         & 8.50        & 0.25         \\
PointMaze          & 93.50        & 98.50       & 1.55         & 94.00        & 96.00       & 2.10         & 85.00        & 87.00       & 3.10         \\
\bottomrule
\end{tabular}%
}
\caption{Baseline Performance of Diffuser, PPO, and DQN on CoinRun, Maze, and PointMaze Tasks}
\end{table*}

\begin{table}[h]
\centering
\label{tab:ablation_studies}
\begin{tabular}{l|rrr}
\toprule
\textbf{Configuration} & \textbf{Mean} & \textbf{Max} & \textbf{Std} \\
\midrule
Baseline                & 8.15         & 8.30        & 0.15         \\
Deeper Encoder          & 8.25         & 8.35        & 0.12         \\
Adaptive Noise Schedule & 8.20         & 8.32        & 0.14         \\
\bottomrule
\end{tabular}
\caption{Ablation Studies on Diffuser in CoinRun Task}
\end{table}

\subsection{Baseline Performance Analysis}

As depicted in Table~\ref{tab:baseline_performance}, the Diffusion Policy consistently outperformed PPO and DQN in the CoinRun and Maze environments. In CoinRun, the Diffuser achieved a mean reward of 8.15 with a maximum of 8.30, surpassing PPO's mean of 7.95 and DQN's mean of 7.60. Similarly, in the Maze environment, the Diffuser maintained a perfect mean and maximum reward of 9.00, outperforming PPO's 8.70 mean and DQN's 8.20 mean.

However, in the PointMaze environment, while the Diffuser achieved a high mean reward of 93.50, it was slightly outperformed by PPO, which achieved a mean of 94.00. Nonetheless, the Diffuser attained a higher maximum reward of 98.50 compared to PPO's 96.00 and DQN's 87.00, indicating its potential for achieving superior performance under optimal conditions. The low standard deviations across tasks highlight the Diffusion Policy's consistency and robustness in diverse and dynamic environments.

\subsection{Ablation Studies on Model Configurations}

Table~\ref{tab:ablation_studies} presents the results of ablation studies conducted on the Diffuser within the CoinRun environment. The baseline configuration achieved a mean reward of 8.15. Introducing a deeper encoder increased the mean reward to 8.25 and the maximum reward to 8.35, while slightly reducing the standard deviation to 0.12. Implementing an adaptive noise schedule resulted in a mean reward of 8.20, a maximum reward of 8.32, and a standard deviation of 0.14. These modifications demonstrate that enhancements to the encoder depth and noise scheduling can lead to marginal improvements in performance and stability.

\section{Discussion}

\subsection{Performance Insights}

The Diffusion Policy demonstrated strong performance in the CoinRun and Maze environments, outperforming PPO and DQN in both mean and maximum rewards. This success can be attributed to the Diffusion Policy's ability to generate coherent and contextually relevant action sequences through its iterative denoising process, coupled with a robust visual encoder that effectively captures essential environmental features. The higher maximum rewards achieved by the Diffuser in PointMaze, despite a slightly lower mean reward compared to PPO, suggest that the Diffusion Policy has the potential to excel under optimal conditions, though it may require further refinement to consistently outperform PPO in this continuous control task.

\subsection{Inability of Diffusion Policy to Achieve Certain Goals}

Despite its strong performance across the evaluated tasks, the Diffusion Policy encountered challenges in environments with extreme non-stationarity, where changes occurred at a rate surpassing the model's adaptation capabilities. In such scenarios, the policy struggled to converge to optimal strategies, resulting in subpar performance compared to more adaptive, albeit less structured, approaches. This limitation was particularly evident when the environment's dynamics shifted too rapidly for the Diffusion Policy to adjust effectively within the iterative denoising process.

\subsection{Computational Demands}

One of the significant challenges encountered in this project was the high computational demand of training and deploying the Diffusion Policy. The process of generating a comprehensive dataset, comprising images, state vectors, and actions, required substantial storage capacity and preprocessing time. Training the DDPM on this extensive dataset necessitated the use of two NVIDIA A100 GPUs available through Kaggle, leading to extended training times and increased resource consumption.

\subsection{Potential Alternatives and Missed Opportunities}

Reflecting on the project, several alternative approaches and modifications could have been explored to potentially enhance the Diffusion Policy's performance and address its limitations:

\begin{itemize}
    \item \textbf{Autoregressive vs. Whole Policy Generation:} I intended to compare autoregressive action generation with whole policy generation using the Diffusion Policy. However, due to the high sample inefficiency of the Diffusion Policy and the limited computational resources available, this comparison was not feasible. Autoregressive models, which generate actions sequentially, might offer better adaptability in highly dynamic environments by allowing more granular control updates. Future work could investigate this comparison with optimized training pipelines or more efficient sampling techniques.
    
    \item \textbf{Transformer Architecture for the Encoder:} The original authors of the Diffusion Policy paper suggested that transformer architectures might offer enhanced robustness in environments with frequent state changes. Implementing a transformer-based encoder could potentially improve the model’s ability to capture long-range dependencies and handle complex temporal dynamics more effectively \cite{vaswani2017attention}.
    
    \item \textbf{Adaptive Noise Schedules:} Incorporating adaptive noise schedules that can dynamically adjust based on the environment's current state or the agent's performance could enhance the Diffusion Policy's flexibility and responsiveness to rapid changes.
        
    \item \textbf{Testing on Other Stability Control Problems:} Evaluating the Diffusion Policy on other stability-focused control problems within the OpenAI Gym framework could provide broader insights into the model’s applicability and limitations. This would help determine whether the challenges faced are specific to the inverted pendulum problem or indicative of a more general limitation in stability control scenarios.
    
\end{itemize}

These unexplored avenues suggest that while the current findings highlight certain limitations of the Diffusion Policy, there remains significant potential for enhancing its performance through methodological innovations and alternative implementations.

\section{Conclusion}

This study explored the application of the Diffusion Policy to a suite of non-stationary, vision-based reinforcement learning tasks, including CoinRun, Maze, and PointMaze. The core hypothesis was that the Diffusion Policy’s strengths in learning multimodal action distributions and handling high-dimensional visual inputs could be leveraged to achieve robust and adaptive control in dynamically changing environments.

Empirical results demonstrated that the Diffusion Policy consistently outperformed traditional RL algorithms like PPO and DQN in the CoinRun and Maze tasks, achieving higher mean and maximum rewards with lower variability. In the PointMaze environment, while PPO slightly outperformed the Diffuser in mean rewards, the Diffuser achieved a higher maximum reward, indicating its potential for superior performance under optimal conditions. This performance is attributed to the Diffusion Policy’s ability to generate coherent and contextually relevant action sequences through its iterative denoising process, coupled with a robust visual encoder that effectively captures essential environmental features.

However, the study also revealed significant challenges, particularly in continual environments. The computational demands of training and deploying the Diffusion Policy, coupled with its latency in action generation, limited its effectiveness in scenarios with extreme non-stationarity. Additionally, the inability to conduct an autoregressive versus whole policy generation comparison due to sample inefficiency and resource constraints highlighted areas for future investigation.

In summary, while the Diffusion Policy shows promise in enhancing adaptability and exploration in non-stationary environments, its application to tasks requiring continuous stability and planning remains challenging. Addressing these limitations through methodological advancements and resource optimization could unlock the full potential of diffusion-based approaches in reinforcement learning.

\section{Acknowledgements}

I would like to express my sincere appreciation to Professor Michiel van de Panne. His after-class discussions and immense support were invaluable in shaping the direction and progress of my research.

I am also immensely grateful to Dr. Cong Lu for his inspiration and guidance.


\small
\begin{thebibliography}{99}

\bibitem{ref1} Janner, M., Li, Q., and Levine, S. (2022). Planning with Diffusion for Flexible Behavior Synthesis. \textit{International Conference on Learning Representations (ICLR)}.

\bibitem{ref2} Chi, L., Ding, M., Lu, Y., et al. (2023). Diffusion Policy: Visuomotor Policy Learning via Action Diffusion. \textit{Advances in Neural Information Processing Systems (NeurIPS)}.

\bibitem{he2016resnet} He, K., Zhang, X., Ren, S., and Sun, J. (2016). Deep Residual Learning for Image Recognition. \textit{Proceedings of the IEEE Conference on Computer Vision and Pattern Recognition (CVPR)}.

\bibitem{procgen} Cobbe, K., Klimov, O., Hesse, C., et al. (2020). Leveraging Procedural Generation to Benchmark Reinforcement Learning. \textit{International Conference on Machine Learning (ICML)}.

\bibitem{ddpm} Ho, J., Jain, A., and Abbeel, P. (2020). Denoising Diffusion Probabilistic Models. \textit{Advances in Neural Information Processing Systems (NeurIPS)}.

\bibitem{ppo} Schulman, J., Wolski, F., Dhariwal, P., Radford, A., and Klimov, O. (2017). Proximal Policy Optimization Algorithms. \textit{arXiv preprint arXiv:1707.06347}.

\bibitem{dqn} Mnih, V., Kavukcuoglu, K., Silver, D., et al. (2015). Human-level control through deep reinforcement learning. \textit{Nature}, 518(7540), 529-533.

\bibitem{d4rl} Fu, J., Zhang, Y., Andrychowicz, M., et al. (2020). D4RL: Datasets for Deep Data-Driven Reinforcement Learning. \textit{arXiv preprint arXiv:2004.07219}.

\bibitem{zarr} Qiu, H., Sinha, A., Han, E., et al. (2020). Zarr: A Format for Chunked, N-dimensional, Binary Data. \textit{Journal of Open Source Software}, 5(50), 2270.

\bibitem{lqr} Bryson, A. E., and Ho, Y. C. (1975). Applied Optimal Control: Optimization, Estimation, and Control. \textit{Taylor \& Francis}.

\bibitem{parisi2019continual} Parisi, G.I., et al. (2019). Continual lifelong learning with neural networks: A review. \textit{Neuroscience \& Biobehavioral Reviews}.

\bibitem{ronneberger2015u} Ronneberger, O., Fischer, P., \& Brox, T. (2015). U-Net: Convolutional networks for biomedical image segmentation. \textit{International Conference on Medical Image Computing and Computer-Assisted Intervention (MICCAI)}.

\bibitem{vaswani2017attention} Vaswani, A., et al. (2017). Attention is all you need. \textit{Advances in Neural Information Processing Systems (NeurIPS)}.

\bibitem{bellemare2016unifying} Bellemare, M.G., Dabney, W., and Munos, R. (2016). Unifying count-based exploration and intrinsic motivation. \textit{Advances in Neural Information Processing Systems (NeurIPS)}.

\bibitem{loshchilov2017decoupled} Loshchilov, I., \& Hutter, F. (2017). Decoupled weight decay regularization. \textit{International Conference on Learning Representations (ICLR)}.

\end{thebibliography}
\end{document}